\documentclass{mva_style}
\usepackage{graphicx}
\usepackage[english]{babel}
\usepackage[nottoc]{tocbibind}
\usepackage{hyperref}
\usepackage{svg}
\usepackage{amsmath}
\usepackage{multirow}

\finalcopy 

\begin{document}
\title{Image Based Review Text Generation with Emotional Guidance}

\author{
  Xuehui Sun\\
  Shanghai Jiao Tong University\\
  {\tt zidaneandmessi@sjtu.edu.cn}\\
  \and
  Zihan Zhou\\
  Shanghai Jiao Tong University\\
  {\tt footoredo@sjtu.edu.cn}\\
  \and
  Yuda Fan\\
  Shanghai Jiao Tong University\\
  {\tt kurodakanbei@sjtu.edu.cn}\\
}

\maketitle

\section*{\centering Abstract}
\textit{
     In the current field of computer vision, automatically generating texts from given images has been a fully worked technique. Up till now, most works of this area focus on image content describing, namely image-captioning. However, rare researches focus on generating product review texts, which is ubiquitous in the online shopping malls and is crucial for online shopping selection and evaluation. Different from content describing, review texts include more subjective information of customers, which may bring difference to the results. Therefore, we aimed at a new field concerning generating review text from customers based on images together with the ratings of online shopping products, which appear as non-image attributes. We made several adjustments to the existing image-captioning model to fit our task, in which we should also take non-image features into consideration. We also did experiments based on our model and get effective primary results.
}

\section{Introduction}

\begin{figure}
  \centering
  \includegraphics[width=\linewidth]{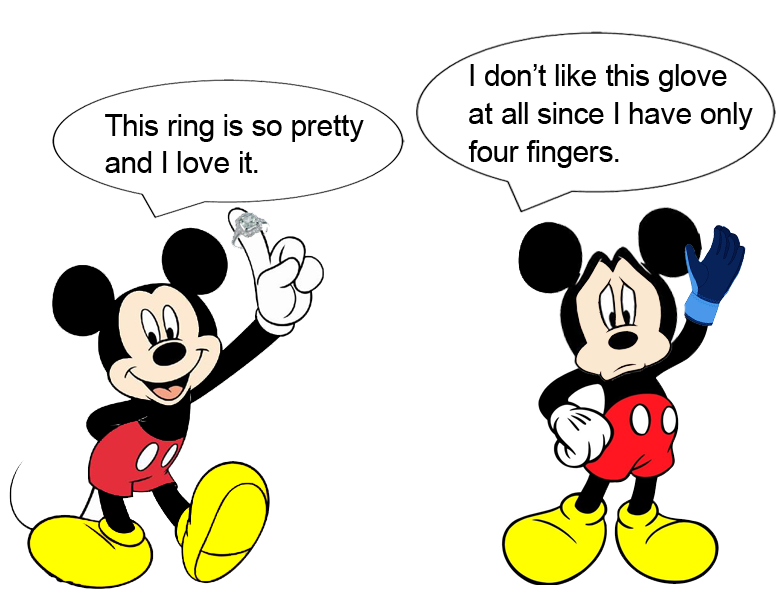}
  \caption {Review texts with different emotions.}
  \label{fig:sample}
\end{figure}

Traditional image-captioning work usually intends to describe the content of a given image using human-like properly formed English sentences~\cite{showandtell, showattendandtell, semanticattention, watchwhatyoujustsaid}. Different from general image-captioning, review texts are expected to contain the evaluation of the product and emotion of the reviewer instead of merely describing the picture. Although the data of real review texts is quite sufficient, few works have taken an insight into this topic. Our purpose is to generate review texts with emotions, describing the possible feedback from those who had bought this product, just like Figure~\ref{fig:sample} presents.

In a sense, our work is quite close to the traditional image-captioning work, and image plays a significant role deciding the content of our resulting texts. That means it is sensible to follow preceding models to achieve our target of review text generating based on images. However, if we only use images alone, the result will be more closer to objective description rather than subjective reviewing, lacking subjective information like orientation and emotion. Therefore, we need to use non-image features to guide our review texts to be generated. Moreover, non-image features may have different structures and properties for perception comparing to image features. To be specific, image features describe concrete vision information, but non-image features characterize abstract information. In consideration of these differences, we need to employ extra skills to make them work together but separately.

We utilize the reviews of products from customers to guide our output to imitate. With product images and other non-image attributes, we can build a caption-like deep learning model. To make sure our review texts contains appropriate emotions, we use the ratings of the reviewers on products to direct the review texts. These ratings are included in the dataset, representing a general evaluation of the purchaser.  Inspired by work based on generating text reviews from aspect-sentiment scores \cite{generatefromscores}, we decide to combine the aspect of images and scores and translate them into review texts.

Image-captioning models mostly follows encoder-decoder architecture, which is first proposed and used by Vinyals et al.~\cite{showandtell}. They use deep convolution neural network (CNN) to extract image features, performing as the encoder, and Recurrent Neural Networks (RNN) like Long Short-Term Memory (LSTM)~\cite{lstm} to convert features to texts, acting as the decoder. To make outputs resemble the real words, we employ Guiding Long-Short Term Memory networks (gLSTM)~\cite{glstm} to generate the words. Meanwhile, utilizing the same framework, we can figure out what are customers interested in by adopting the attention mechanism proposed by \cite{showattendandtell}. Different from image-captioning tasks where the text directly describes the image, the connection between image and text is not that tight in our settings. Imagine doing both image-captioning and review text generation with the same picture, say a ring with a diamond. In an image-captioning task, simply ``A ring with a diamond'' would be sufficient. But in a review text generation task, a more commonsensical solution may be ``I like the ring for the beautiful diamond in it.'' In a sense, in each step of generation, we have to \emph{think} about what we are going to focus on, instead of just focusing on the next describable object. This is why the attention mechanism in \cite{watchwhatyoujustsaid} which directly generates attention mask from the previous word is not applicable here. To solve this issue, we designed a bilevel gLSTM model, in which a new lower-level gLSTM is utilized to generate the current context information, which the attention mask is based upon. The higher-level gLSTM is a Time-Dependent gLSTM (td-gLSTM) which is introduced in \cite{watchwhatyoujustsaid}, except there is now an additional field in the guidance for non-image feature.

In our task, both image features and non-image features are essential and each plays a different role. The image features provide the entities (watch, ring, shoes, etc.) and attributes (color, texture, shape, etc.) to review with, or in other words, \emph{what to review}. On the contrary, non-image features determine the background setting and the underlying tone of the review, or \emph{how to review}. In a sense, non-image features are more like static global guidance throughout the generation process, while image features are directly used to determine what to say in the next step of generation. This contradiction is why we adopt the attention mechanism for the image features and gLSTM model for the non-image features.

\section{Related Work}

Since the early success of applying deep neural networks in image-captioning tasks such as~\cite{Karpathy_2015_CVPR,showandtell} , numerous methods have been developed to solve this problem. Vinyals et al.~\cite{showandtell} proposes an encoder-decoder framework, using CNN to extract image features, and LSTM~\cite{lstm} to generate description texts. This structure implements an end-to-end model, maximizing the likelihood of the target description sentence. To fix problems such as image information losing for more descriptive captions and regional features over-balancing, in recent researches, attention mechanism has been widely adopted since it was first introduced by Xu et al.~\cite{showattendandtell}. This mechanism aims to make salient features take precedence over other ones. In \cite{showattendandtell}, two different types of attention were introduced, namely ``Stochastic Hard Attention'' and ``Deterministic Soft Attention''. You et al.~\cite{semanticattention} combines bottom-up concepts (visual attributes extracted by different possible methods) with top-down attributes from the CNN model, and uses ``input and output attention'' to implement a real-time generating word prediction. The attention mechanism utilizes ample visual semantic aspects of local parts besides natural global information of images. Zhou et al.~\cite{watchwhatyoujustsaid} raises a new method called ``Text-Conditional Attention'', which uses context information to guide while generating attention masks.

Jia et al.~\cite{glstm} introduces an extended LSTM model called Guiding Long-Short Term Memory network (gLSTM). It adds semantic information extracted from the images as an extra input to each LSTM block in order to guide the model, making the result more connected to the content of images. And Zhou et al.~\cite{watchwhatyoujustsaid} raises a further enhanced Time-Dependent gLSTM (td-gLSTM) model, which changes the property of time-invariant guidance in gLSTM, and allows the guidance to evolve over time. We draw on the experience of these works and also use the td-gLSTM model for our text generating.

In the field of product reviews generation, Zhang et al.~\cite{generatefromscores} uses aspect-sentiment product scores to generate review texts aligned with the aspects and ratings. It proposes Sequential Review Generation Models (SRGMs) and Hierarchical Review Generation Models (HRGMs) to implement the generation. However, the target of this model is concentrated, aiming at generating reviews at a single aspect of vehicles. Therefore, this model does not use any visual information of the products. For our goal, both image and non-image information is critical. It may bring monotonicity or repeatability to the resulting review text without either of them. Based upon existing image-captioning and review texts generating models, we use both image and non-image features. That is where the innovation of our work is.

\section{Approach}

\begin{figure*}
  \centering
  \includegraphics[width=\linewidth]{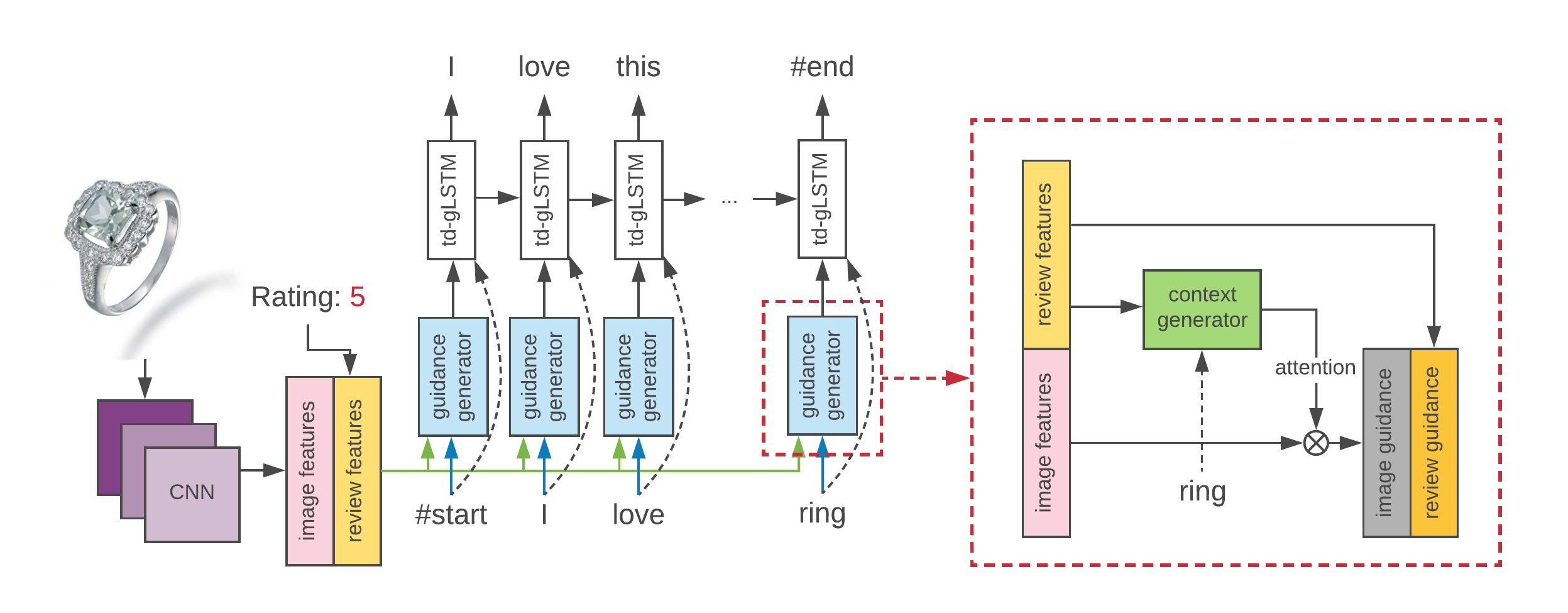}
  \caption {The sketch graph of our model. The graph on the left is an overview and the graph on the right provides detailed structure of every guidance generator.}
  \label{fig:model}
\end{figure*}

There are three parts of input needed: product image features pre-extracted by CNN, given non-image features of products and word-embedding vectors converted from currently generated review texts by a pre-trained word-embedding model.

Figure~\ref{fig:model} shows an overview of our model using bilevel gLSTM. The lower-level gLSTM uses embedding vector of already generated words as its input and gives out an attention mask. The higher-level td-gLSTM receives different features in conjunction and generates the next word.

\subsection{Raw Image Feature Encoder}

The first part of our network is the raw image feature encoder, which extracts the features of product images without any other processing methods. A simple deep CNN model is used to convert given images into image feature vectors. In the model training procedure, this is only used in end-to-end training. Therefore, this part can be omitted when the training dataset contains pre-extracted image features. The result of raw image feature encoder is:

\begin{equation}
f=\mathcal{F}(a),
\end{equation}

where $a$ is the input product image, f is the image feature vector, and $\mathcal{F}$ denotes CNN encoding function.

\subsection{Guidance Generator}

The hardcore of our model is the guidance generator. This ``guidance'' here denotes the guidance to the td-gLSTM block, which consists of image features and non-image features, as two different types of guidance. The most critical part of our guidance generator is the generating of image guidance.

We apply the attention mechanism from \cite{showattendandtell} in order to make our network ``pay attention to'' certain definitive areas of input images, guided by context information. We use the image features obtained above as inputs, and generate an attention mask to automatically locate the area we want to focus on in the feature maps.

Inspired by ``Text-Conditional Attention'' method introduced in \cite{watchwhatyoujustsaid}, we use word-embedding vectors for semantic guidance to generate attention masks. Nevertheless, different from the original text-conditional attention method, we use a gLSTM model here. This gLSTM receives a word-embedding vector of already generated words and outputs the corresponding attention mask at the current time step. It uses non-image features as guidance. Non-image features denote the ratings of products given by purchasers, which appears in the dataset.  According to the gLSTM model, we define the memory cell and gates in each gLSTM block as follows:
\begin{align}
& i_t = \sigma(W_{ix}x_t+W_{im}m_{t-1}+W_{iq}g_t) \nonumber \\
& f_t = \sigma(W_{fx}x_t+W_{fm}m_{t-1}+W_{fq}g_t) \nonumber \\
& o_t = \sigma(W_{ox}x_t+W_{om}m_{t-1}+W_{oq}g_t) \nonumber \\
& c_t = f_t\odot c_{t-1}+i_t\odot\textup{tanh}(W_{cx}x_t+W_{cm}m_{t-1}+W_{cq}g_t) \nonumber \\
& m_t = o_t\odot c_t.
\end{align}
Here $i_t, f_t, o_t, c_t$ are the input gate, the forget gate, the output gate, and the memory cell of the LSTM model at time state $t$. $m_t$ is the hidden state, and also the output of the current block generated by the memory cell, appearing as the target attention mask here. $x_t$ is the input texts, which is the word-embedding vector converted from input words already generated from the result review texts. $W_{[\cdot][\cdot]}$ denotes the weight parameters that need to be trained. $\sigma(\cdot)$ and $\textup{tanh}(\cdot)$ are activation functions of sigmoid and hyperbolic. $\odot$ represents element-wise multiplication. $g_t$ denotes the guidance information, which is actually non-image feature vectors here. We use non-image features to guide our gLSTM model to generate attention masks from already generated words. We can write the result as a more simplified version:

\begin{equation}
m_t=\mathcal{G}(x_t, g_t),
\end{equation}

where $\mathcal{G}(x_t, g_t)$ is the gLSTM decoding function with $x_t$ as input and $g_t$ as guidance. The guidance indeed means using given product ratings to decide the underlying emotion of our target review texts. Then the image feature vectors obtained by raw image feature encoder will be encoded again by this attention mask. Finally, we get attention image feature vectors, which is the result of the attention mechanism.

\newcommand{\tabincell}[2]{\begin{tabular}{@{}#1@{}}#2\end{tabular}}
\begin{table*}[ht]
 \centering
\caption{\label{tab:test}Sample test results of our experiments}
\renewcommand\arraystretch{1.2}
 \begin{tabular}{|c|c|l|}
  \hline
 \tabincell{c}{Product \\ image} & Rating & \multicolumn{1}{c|}{Generated review texts}\\
  \hline

\multirow{3}{*}{\includegraphics[width=0.5in]{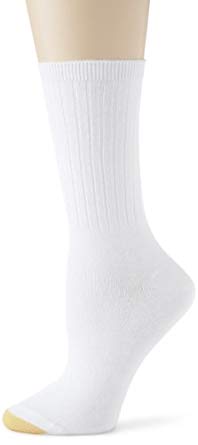}} & 5.0 & \tabincell{l}{I love these socks. They are very comfortable and fit well and are very a great of the socks.\\ They are very comfortable and I would recommend to anyone looking for a nice pair of \\socks...} \\
\cline{2-3}&
 4.0 &\tabincell{l}{I like these. They are a little stiff but not too. I love the way they look and but they are a\\ little tight on my feet. I was not sure that they would be a little bit too but for the price I\\ would have ordered a size larger than I usually wear socks...}\\
\cline{2-3}&
 1.0 &\tabincell{l}{I don't like the socks. I had to return them. I ordered a size large and these were not. \\They fit like regular size and they are very thin. The socks didn't shrink in...}\\
\hline

\multirow{2}{*}{\includegraphics[width=0.5in]{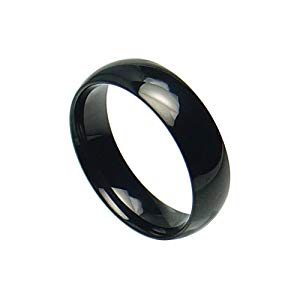}} & 5.0 & \tabincell{l}{I bought this ring to my husband for Christmas. I was very pleased with it. This piece of \\this ring is a little but it was a great. I am very happy with this product...} \\
\cline{2-3}&
 1.0 &\tabincell{l}{I really hope it as much as the same picture but I was not very happy with the ring that I \\ was looking for. This is the first one i bought and I was very disappointed in this \\piece of this type...}\\
 \hline
\multirow{3}{*}[-5pt]{\includegraphics[width=0.5in]{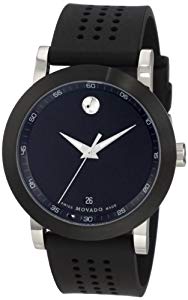}} & 5.0 & \tabincell{l}{I bought this watch for my boyfriend and he loves it.  The only problem I have with it it \\was  little  bit too long.  I was very happy with the quality of this watch...} \\
\cline{2-3}&
 3.0 &\tabincell{l}{I was a little disappointed about this watch. It is a little hard to find the watch for the \\ price. I received this a new one and the is very...}\\
\cline{2-3}&
 1.0 &\tabincell{l}{This is the worst product I bought from Amazon. I have been wearing it for a few days \\and I was able to get the bracelet. The band is too small...}\\
\hline

\multirow{3}{*}[-20pt]{\includegraphics[width=0.5in]{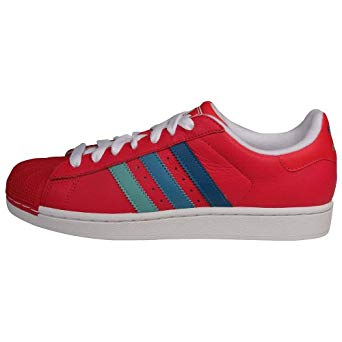}} & 5.0 & \tabincell{l}{I love these shoes. I have been very high and the lining is very good and i have no \\complaints about them and it fit me perfectly...} \\
\cline{2-3}&
 3.0 &\tabincell{l}{These shoes are cute, and for a long period of time I have a wide foot, and they are still in\\ good quality. I returned it for the perfect size. The color is way too tight...}\\
\cline{2-3}&
 1.0 &\tabincell{l}{I have not had the same pair of. This one is a size too small. The material of the sole is \\very stiff and uncomfortable. I would not buy this... }\\
\hline

\multirow{2}{*}{\includegraphics[width=0.5in]{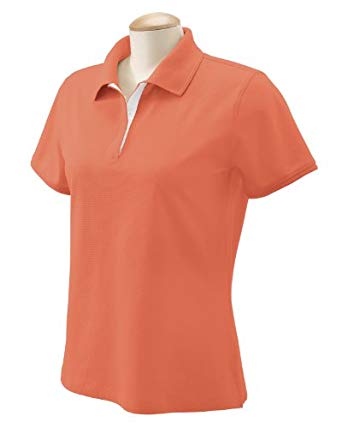}} & 5.0 & \tabincell{l}{I bought this shirt for my son. They are very soft and the fit is perfect and the fit is true \\to size. I would recommend this to anyone who wants to wear a t-shirt or a shirt to use...} \\
\cline{2-3}&
 2.0 &\tabincell{l}{It was not the same as the picture and the material is not good. I ordered a small size and \\I ordered a size larger than I thought it would be. It doesn't fit a bit. I normally wear \\a size 8...
}\\
\hline
 \end{tabular}
  \label{tab}
\end{table*}

After applying the attention mask to the raw image feature vectors, we finish the image guidance generating process. Next, we concatenate the image guidance vector with the review guidance. We use non-image features as review guidance, namely the ratings of products in the dataset mentioned above. The result of guidance generating is:

\begin{equation}
\hat{g}_t=(f\odot m_t)\oplus g_t,
\end{equation}

where $\hat{g}_t$ is the combined guidance (feature) vector, $f\odot m_t$ is image feature vector with attention mask applying. $\oplus$ denotes vector concatenating operation.

\subsection{Feature Decoder}

The last part of our model decodes all features and generates the target output words. After concatenating both feature vectors above, the resulting vector is used as the time-dependent guidance of a td-gLSTM model. In every unit of time, the above models are repeated, and a new word of the output sequence is generated. Our final result is:

\begin{equation}
y_t=\mathcal{D}(x_t, \hat{g}_t),
\end{equation}

where $\mathcal{D}(x_t, \hat{g}_t)$ is the td-gLSTM decoding function with $x_t$ as input and $\hat{g}_t$ as guidance. $y_t$ is the next word we obtain.




\balance

\section{Experiments}

\subsection{Dataset}

Our experiment is based on Amazon product data introduced in \cite{mcauley2015}. This dataset contains real-world product information and user reviews on \href{https://www.amazon.com/}{Amazon.com} over the past decades, in which the product images and user reviews are what we need.

This dataset includes millions of products from 24 different categories. In our experiment, we only focus on the ``Clothing, Shoes and Jewelry'' category. The review texts are English sentences, and the reviewers' ratings are integers from 1 to 5.

\subsection{Visual features}

We use the features presented in \cite{he2016}, which is extracted using a pre-trained convolutional neural network, the Caffe reference model \cite{jia2014caffe}. This model has been shown to be useful for this type of images in \cite{mcauley2015}
 and \cite{he2016}.  It implements the architecture in \cite{imagenet} with 5 convolutional layers followed by 3 fully-connected layers and was pre-trained on 1.2 million ImageNet (ILSVRC2010) images. Each extracted feature is taken from the output of its second fully-connected layer (FC7) with length $F=4096$.

\subsection{Performance}

We did our experiment using a trained model based on $40000$ sets of data including image feature vectors, non-image ratings, and ground-truth review texts. We filtered those data with review text longer than $100$ out, and $34810$ sets of data remained. The size of word-embedding dictionary is $2934$. The output text also has length $100$.

Table~\ref{tab} shows some sample test results of our experiments. It turns out that out model can successfully generate review texts related with correct product information and the emotion of generated texts can change with different review ratings.  Our training dataset is relatively small due to time and source limiting, so the result still faces some problems occasionally such as repetition of words, misidentification and generating meaningless texts. But definitely our model will work better with larger datasets or more skillful natural language processing methods. We also did ablation experiments of replacing the gLSTM-based attention by ``Deterministic Soft Attention" method in \cite{showattendandtell}. But due to the insufficiency of our dataset, the results did not show too much defects, and the superiority of the final model using gLSTM is not that distinguishable from words. For this reason, we did not display the results of ablation experiments, but only displayed the results of our final models from different products and different ratings.

\bibliographystyle{ieeetr}
\bibliography{sample}

\begin{thebibliography}{10}

\bibitem{showandtell}
O.~Vinyals, A.~Toshev, S.~Bengio, and D.~Erhan, ``Show and tell: A neural image
  caption generator,'' in {\em CVPR}, 2015.

\bibitem{showattendandtell}
K.~Xu, J.~L. Ba, R.~Kiros, K.~Cho, A.~Courville, R.~Salakhutdinov, R.~S. Zemel,
  and Y.~Bengio, ``Show, attend and tell: Neural image caption generation with
  visual attention,'' {\em arXiv preprint arXiv:1502.03044}, 2015.

\bibitem{semanticattention}
Q.~You, H.~Jin, Z.~Wang, C.~Fang, and J.~Luo, ``Image captioning with semantic
  attention,'' in {\em CVPR}, 2016.

\bibitem{watchwhatyoujustsaid}
L.~Zhou, C.~Xu, P.~Koch, and J.~J. Corso, ``Watch what you just said: Image
  captioning with text-conditional attention,'' {\em arXiv:1606.04621}, 2016.

\bibitem{generatefromscores}
H.~Zhang and X.~Wan, ``Towards automatic generation of product reviews from
  aspect-sentiment scores,'' in {\em INLG}, 2017.

\bibitem{lstm}
S.~Hochreiter and J.~Schmidhuber, ``Long short-term memory. neural
  computation,'' {\em 9(8):1735–1780.}, 1997.

\bibitem{glstm}
X.~Jia, E.~Gavves, B.~Fernando, and T.~Tuytelaars, ``Guiding long-short term
  memory for image caption generation,'' {\em arXiv:1509.04942}, 2016.

\bibitem{Karpathy_2015_CVPR}
A.~Karpathy and L.~Fei-Fei, ``Deep visual-semantic alignments for generating
  image descriptions,'' in {\em The IEEE Conference on Computer Vision and
  Pattern Recognition (CVPR)}, June 2015.

\bibitem{mcauley2015}
J.~{McAuley}, C.~{Targett}, Q.~{Shi}, and A.~{van den Hengel}, ``Image-based
  recommendations on styles and substitutes,'' in {\em SIGIR}, 2015.

\bibitem{he2016}
R.~{He} and J.~{McAuley}, ``Ups and downs: Modeling the visual evolution of
  fashion trends with one-class collaborative filtering,'' in {\em WWW}, 2016.

\bibitem{jia2014caffe}
Y.~Jia, E.~Shelhamer, J.~Donahue, S.~Karayev, J.~Long, R.~Girshick,
  S.~Guadarrama, and T.~Darrell, ``Caffe: Convolutional architecture for fast
  feature embedding,'' {\em arXiv preprint arXiv:1408.5093}, 2014.

\bibitem{imagenet}
A.~Krizhevsky, A.~Krizhevsky, and G.~E. Hinton, ``Imagenet classification with
  deep convolutional neural networks,'' in {\em NIPS}, 2012.

\end{thebibliography}

\end{document}